\documentclass[sigconf,nonacm]{acmart}

\AtBeginDocument{%
  }

\usepackage{booktabs}
\usepackage{subcaption}
\usepackage{amsmath}

\usepackage{amssymb}
\usepackage{algorithm}
\usepackage{algorithmic}
\usepackage{multirow}
\usepackage{relsize}
\usepackage{xcolor}
\usepackage{graphicx}
\usepackage{tikz}
\usetikzlibrary{calc,positioning,arrows.meta,fit}

\usepackage{hyperref}
\usepackage[capitalize,noabbrev]{cleveref}
\crefname{appendix}{Appendix}{Appendices}
\Crefname{appendix}{Appendix}{Appendices}

\newcommand{\method}{{\smaller HUVR+SIREN}}

\begin{document}
\title{Rethinking Amortized Neural Representations for High-Resolution Terrain Elevation Data}

\author{Haoan Feng}
\affiliation{%
  \institution{University of Maryland, College Park}
  \state{MD}
  \country{USA}
}
\email{hfengac@umd.edu}

\author{Xin Xu}
\affiliation{%
  \institution{University of Maryland, College Park}
  \state{MD}
  \country{USA}
}
\email{xinxu629@umd.edu}

\author{Leila De Floriani}
\affiliation{%
  \institution{University of Maryland, College Park}
  \state{MD}
  \country{USA}
}
\email{deflo@umd.edu}

\renewcommand{\shortauthors}{Feng et al.}

\begin{abstract}
Implicit neural representations (INRs) model a signal as a continuous coordinate-to-value function. For terrain elevation data, this supports analytic derivatives, arbitrary-resolution decoding, and a smooth surface model of the underlying heightfield. However, fitting and storing a separate INR for every tile does not scale to large terrain datasets. Amortized neural representations reduce this cost with a shared network: a new tile is mapped to a compact per-tile payload, and a shared decoder reconstructs the heightfield from it. Most such methods are hypernetworks that predict the payload in a single forward pass, while others recover it through a short per-tile optimization. These methods were developed primarily for natural images, and their suitability for terrain heightfields remains unclear. We introduce a controlled benchmark on a $1\,\text{m/pixel}$ terrain dataset and evaluate three representative methods under a unified protocol. Observing a clear cross-domain gap, we propose \emph{HUVR+SIREN}, a hypernetwork that adapts the strongest benchmarked method (HUVR) by replacing its coordinate decoder with a smooth, analytically differentiable one. It attains the best height and derivative fidelity on the benchmark with no additional per-tile storage and lower decode cost, and tolerates aggressive post-training quantization with negligible quality loss, giving a compact terrain neural format. Ablations and diagnostics further identify which design choices transfer to terrain and show that the per-tile bottleneck is already near its useful limit, leaving the remaining gap in the shared hypernetwork's architectural design.
\end{abstract}

\begin{CCSXML}
<ccs2012>
   <concept>
       <concept_id>10010147.10010257.10010293.10010294</concept_id>
       <concept_desc>Computing methodologies~Neural networks</concept_desc>
       <concept_significance>500</concept_significance>
       </concept>
   <concept>
       <concept_id>10010147.10010371.10010396</concept_id>
       <concept_desc>Computing methodologies~Shape modeling</concept_desc>
       <concept_significance>300</concept_significance>
       </concept>
   <concept>
       <concept_id>10010147.10010371.10010395</concept_id>
       <concept_desc>Computing methodologies~Image compression</concept_desc>
       <concept_significance>100</concept_significance>
       </concept>
   <concept>
       <concept_id>10002951.10003227.10003236.10003237</concept_id>
       <concept_desc>Information systems~Geographic information systems</concept_desc>
       <concept_significance>300</concept_significance>
       </concept>
 </ccs2012>
\end{CCSXML}

\ccsdesc[500]{Computing methodologies~Neural networks}
\ccsdesc[300]{Computing methodologies~Shape modeling}
\ccsdesc[100]{Computing methodologies~Image compression}
\ccsdesc[300]{Information systems~Geographic information systems}

\keywords{Implicit neural representations, Digital elevation model, Amortized neural network, Hypernetwork}

\maketitle

\section{Introduction}
\label{sec:intro}

High-resolution digital elevation models (DEMs) are increasingly
captured at sub-meter ground sampling by airborne and satellite
programs, and they sit at the center of much downstream geospatial
analysis, from slope and curvature estimation in geomorphology to
hydrological flow modeling and visibility computation. A growing
body of work treats such a terrain heightfield not as a discrete
raster but as a continuous implicit surface, represented by a small
coordinate network whose analytic derivatives can be evaluated at
arbitrary resolutions and used directly for downstream geomorphological
analysis~\cite{feng2024implicitterrain}.

Implicit neural representations (INRs)~\cite{sitzmann2020siren,
mildenhall2021nerf} provide a compact way to encode this
continuous-surface view. An INR represents a terrain tile as a neural
function that maps a continuous coordinate to elevation. The fitted
network can be queried at arbitrary locations, supports analytic
derivatives, and represents the surface through learned parameters
rather than fixed raster samples. The practical obstacle is that
classical INR fitting trains a new network from scratch for every
tile, which scales poorly to large DEM datasets. Fitting a single
tile to good fidelity takes minutes of per-tile optimization on a GPU,
whereas a trained amortized hypernetwork produces a comparable per-tile
representation in a single forward pass, on the order of milliseconds.

\begin{figure*}[t]
  \centering
  \includegraphics[width=\linewidth]{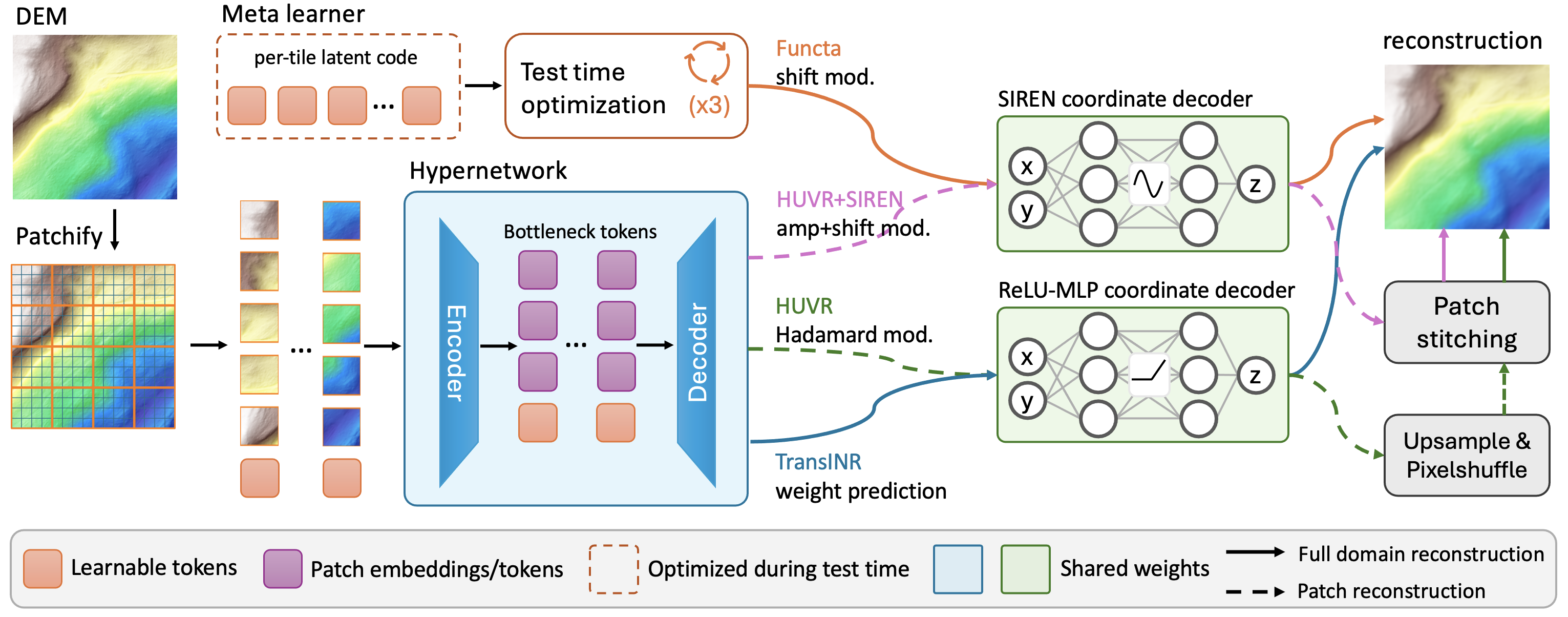}
  \vspace{-0.6cm}
  \caption{Unified view of the four amortized neural representations
    compared. TransINR, HUVR, and \method{} share a
    Transformer-based hypernetwork that conditions a neural coordinate decoder,
    and differ in the conditioning signal, the decoder family, and
    the reconstruction domain. Functa bypasses the hypernetwork; its per-tile
    latent code is optimized at test time by a short meta-learner
    inner loop and directly conditions a SIREN neural coordinate decoder.}
  \Description[Unified architecture diagram of four amortized
    neural representations]{A wide schematic showing a shared
    spine that connects an input digital elevation tile through a
    hypernetwork (Transformer encoder, intermediate bottleneck
    tokens, Transformer decoder) and a small neural coordinate
    decoder that maps a coordinate (x, y) to a height z. Four
    colored branches off this spine indicate the four methods:
    TransINR predicts the full ReLU-MLP weights; HUVR produces
    Hadamard amplitude modulation feeding a ReLU MLP plus
    upsample-and-pixelshuffle decoder; \method{} produces
    amplitude-plus-shift modulation feeding a SIREN coordinate
    decoder; Functa bypasses the hypernetwork and instead
    optimizes a per-tile latent code at test time via a
    meta-learner inner loop that directly conditions the shared
    SIREN. Patch-based methods are drawn with a dashed path that
    converges on a patch-stitching block; the full-domain TransINR
    path is drawn solid.}
  \label{fig:arch}
\end{figure*}

\emph{Amortized} neural representations are a candidate remedy for
this scaling cost. A shared model is trained once across the dataset
so that each new tile is represented by a small per-tile token
payload, and a shared coordinate decoder reconstructs the signal from
that token. Most such methods are hypernetworks that predict the
payload in a single forward pass. A second family instead recovers it
through a short per-tile optimization at inference (\cref{fig:arch}).
Either way, the stored or transmitted artifact is the per-tile token
payload rather than a full per-tile network. These constructions have
been studied primarily on natural images. Their behavior on high-resolution terrain heightfields
remains unclear, because terrain differs strongly from RGB images in
channel structure, smoothness, and derivative requirements. This
paper addresses that cross-domain adaptation problem: whether
amortized neural representations can serve as effective terrain data
formats, and which design choices are needed for them to transfer. Our main contributions are threefold:
\begin{itemize}
  \item We benchmark three representative amortized-INR methods
    (TransINR~\cite{chen2022transinr}, Functa~\cite{dupont2022functa},
    and HUVR~\cite{gwilliam2026huvr}) on a high-resolution ($1\,\text{m/pixel}$) swisstopo
    terrain dataset under a unified protocol, characterizing each across
    height fidelity, gradient and Laplacian fidelity, decode cost, and storage cost
    under post-training quantization. To our knowledge
    this is the first controlled comparison of amortized INRs on
    high-resolution terrain DEMs.
  \item Motivated by the overall smoothness of terrain surfaces and the
    derivative fidelity requirements of downstream geomorphological
    analysis~\cite{feng2024implicitterrain}, we introduce \method{},
    a bounded adaptation that replaces the ReLU coordinate decoder of
    the strongest benchmarked method (HUVR) with SIREN~\cite{sitzmann2020siren}.
    This terrain data adaptation raises fitting accuracy (PSNR $+2.83\,\text{dB}$)
    at identical per-tile storage, with consistent
    gains in derivative fidelity, and yields a post-training quantization frontier
    that dominates HUVR's full-precision (fp32) configuration.
  \item To locate the source of the residual quality gap, we combine
    ablation studies with two dedicated diagnostics on \method{}, a
    per-instance fitting bound and a per-tile full-model upper bound.
    These analyses localize the performance gap to the shared post-bottleneck pipeline,
    indicating that closing it will likely require improving the shared
    hypernetwork architecture rather than enlarging the per-tile token capacity.
\end{itemize}

\section{Related Work}
\label{sec:related-work}

\subsection{Implicit neural representations and amortized neural representations}
\label{sec:related:amortized}

Implicit neural representations encode a signal as a coordinate
network whose weights parameterize the signal itself. The foundational
constructions span three signal domains: signed distance fields
(DeepSDF~\cite{park2019deepsdf}) and occupancy
fields~\cite{mescheder2019occupancy} for shape, scene representation
networks (SRN~\cite{sitzmann2019srn}) and neural radiance fields
(NeRF~\cite{mildenhall2021nerf}) for 3D scenes, and sinusoidal-activation
networks (SIREN~\cite{sitzmann2020siren}) for general signals with
analytic derivatives. Two architectural ingredients enable these
networks to fit high-frequency content from low-dimensional coordinate
inputs: input-side Fourier-feature
encodings~\cite{tancik2020fourier} and SIREN's sinusoidal first
layer~\cite{sitzmann2020siren}. INR fitting is per-instance: each
new signal requires its own optimization, which becomes the
bottleneck that this paper's setting aims to remove.

Two approaches amortize this per-instance fit. The first is the
hypernetwork~\cite{ha2017hypernetworks}, in which one network generates
the weights of another. Applied to INRs, a single encoder forward
pass replaces gradient-based fitting, predicting either the full
coordinate-network weights or a low-rank modulation of a shared
coordinate network~\cite{chen2022transinr, kim2023generalizable}. The
second is gradient-based meta-learning~\cite{finn2017maml}, in which a
learned initialization is adapted to each new signal by a small
inner-loop optimization at inference time.
Tancik et al.~\cite{tancik2021learnedinit} applied this template
directly to coordinate-based INRs and showed that meta-learned
initializations converge to high-fidelity reconstructions in tens of
inner-loop steps. Although the two approaches obtain the per-instance
payload differently (a forward pass versus an inner loop), both can
target the same payload form, a feature-wise affine (FiLM)
modulation~\cite{perez2018film,mehta2021modulated} of a shared
coordinate network's activations.

Amortized-INR architectures combine these ingredients into a small
set of named structural components: a \emph{Transformer-based
hypernetwork}~\cite{chen2022transinr} that ingests an input signal
and produces, in one forward pass, the per-instance conditioning of
a shared \emph{neural coordinate decoder} queried at spatial
coordinates. TransINR~\cite{chen2022transinr} realizes the
hypernetwork monolithically and predicts the full weights of a ReLU MLP
decoder; Instance Pattern Composers~\cite{kim2023generalizable} keep
the monolithic transformer but produce a small set of per-instance
content tokens that modulate a shared pattern-composer MLP, an early
move toward the FiLM-style conditioning route.
HUVR~\cite{gwilliam2026huvr} factors it into a Vision
Transformer~\cite{dosovitskiy2020vit} encoder and a separate
transformer decoder around a low-dimensional patch-token bottleneck,
with DINOv3 distillation~\cite{simeoni2025dinov3} tapped at both
endpoints; the decoder outputs FiLM-style~\cite{perez2018film,mehta2021modulated}
modulations of a shared ReLU MLP decoder. Functa~\cite{dupont2022functa}
takes a different route: it removes the hypernetwork and instead uses
a meta-learning strategy in which a meta-learned initialization and
a shared SIREN decoder reach each per-tile modulation by a short
inner loop on the per-tile reconstruction loss, recovering the
per-tile token at test time rather than by a forward pass. Spatial
Functa~\cite{bauer2023spatial} keeps the inner loop but replaces the
global modulation vector with a 2D grid of latents. We take the
global-latent Functa as this family's representative in our benchmark. These methods are the
set we port to terrain heightfields in
\cref{sec:experiments:crossmethod}; \cref{sec:method:inventory}
gives the implementation-level account.

\subsection{Learned representations of terrain heightfields}
\label{sec:related:terrain}

Treating terrain as a continuous implicit surface, rather than as a
discrete raster, is comparatively recent in the geospatial
literature. ImplicitTerrain~\cite{feng2024implicitterrain} fits a
per-tile SIREN to high-resolution heightfields and shows that the
analytic gradient and Laplacian of the fitted network track
raster-derived slope and curvature references closely enough to
support downstream geomorphological extraction, which establishes
derivative fidelity rather than only pixel PSNR as a relevant
evaluation criterion. Neural Elevation
Models~\cite{dai2024neuralelevation} push the per-tile fitting idea
further and report favorable rate-distortion trade-offs on
aerial DEM. A related but distinct strand targets resolution
increase rather than fixed-resolution storage: DEM
super-resolution~\cite{he2022dem} and continuous DEM
upsampling~\cite{yao2024continuous} both build on INR-style decoders
without amortizing across a dataset. None of these prior
constructions amortizes the per-tile fitting cost in the sense used
in this paper.

\subsection{Compact DEM storage and the INR-compression bridge}
\label{sec:related:storage}

Compact DEM storage has a long history, including raster-side codecs
such as LERC~\cite{esri2021lerc}, COG~\cite{ogc2023cog}, and
ZFP~\cite{lindstrom2014zfp}, slope-preserving DEM quantization
methods~\cite{xie2010slope}, and quadtree- and level-of-detail-based
terrain representations~\cite{lindstrom1996realtime,guerin2017interactive}.
These methods define important production baselines for terrain data
formats. In this paper we focus on a complementary question: whether
amortized-INR architectures developed for natural images can transfer
to high-resolution terrain heightfields. We therefore compare against
reproduced amortized-INR baselines in
\cref{sec:experiments:crossmethod} and a per-tile INR fitting
ceiling in \cref{sec:experiments:diagnosis}. A direct rate-distortion
comparison with production DEM codecs is left for future work.

A related neural-compression line represents each signal by fitting
and quantizing its own INR. COIN~\cite{dupont2021coin},
COIN++~\cite{dupont2022coinpp}, Str{\"u}mpler et
al.~\cite{strumpler2022inrcompression}, and
Cool-chic~\cite{ladune2023coolchic} follow this per-instance
compression setting, where the stored artifact is typically the
quantized network or its weights. Our setting is amortized rather
than per-instance: a shared encoder-decoder is trained across tiles,
and each tile is represented only by a compact bottleneck token
payload. Accordingly, the quantization analysis in
\cref{sec:experiments:crossmethod} acts on the per-tile token
payload rather than on per-instance network weights.

\section{Method}
\label{sec:method}

\subsection{Amortized INRs: a coordinate-decoder formulation}
\label{sec:method:formulation}

An implicit neural representation encodes a single signal as a
coordinate network $f_\theta:[-1,1]^2 \to \mathbb{R}^d$ that maps each
spatial coordinate $\mathbf{x}\in[-1,1]^2$ to the corresponding signal
value $f_\theta(\mathbf{x})$. For a heightfield $\mathbf{S}^{(i)} \in \mathbb{R}^{H\times W}$
viewed as discrete samples of an underlying continuous surface
$s^{(i)}:[-1,1]^2 \to \mathbb{R}^d$, per-instance fitting solves
\begin{equation}
  \theta^{(i)} \;=\; \arg\min_{\theta}\; \mathbb{E}_{\mathbf{x}}\!\left[\,L\!\left(f_\theta(\mathbf{x}),\, s^{(i)}(\mathbf{x})\right)\,\right]
  \label{eq:per-instance}
\end{equation}
independently for each tile $i$ under a per-coordinate reconstruction
loss $L$ (mean-squared error throughout this paper). Terrain has
$d{=}1$; $d{=}3$ recovers the natural-image setting of TransINR and
Functa. This per-instance recipe requires a separate optimization and
a separate stored network for every tile, which scales poorly to
dataset-level use.

\emph{Amortized} neural representations remove this cost by sharing a
pipeline across the dataset (\cref{fig:arch}). A
\emph{Transformer-based hypernetwork} (the term introduced by
TransINR~\cite{chen2022transinr}) maps an input tile
$\mathbf{S}^{(i)}$ to a per-tile \emph{bottleneck token} $\zeta^{(i)}$.
Both TransINR and HUVR~\cite{gwilliam2026huvr} realize this
hypernetwork as a Vision Transformer~\cite{dosovitskiy2020vit} that
ingests the tile as image patches. We write the conversion as a
transformation $\zeta^{(i)} = T_\Phi(\mathbf{S}^{(i)})$ with shared
parameters $\Phi$. A shared \emph{neural
coordinate decoder} $f_{\bar\theta}$, a small network with weights
$\bar\theta$ shared across the dataset, is queried at a coordinate
$\mathbf{x}$ and conditioned on $\zeta^{(i)}$ to return the
predicted signal value, so the full per-tile reconstruction is
$f^{(i)}(\mathbf{x}) = f(\mathbf{x};\,\bar\theta,\,\zeta^{(i)})$. The
bottleneck token $\zeta^{(i)}$ is the only per-tile object stored, and
its dimensionality sets the per-tile storage budget.

The hypernetwork parameters $\Phi$ and the shared decoder weights
$\bar\theta$ are trained once across the dataset by jointly minimizing
\begin{equation}
  \min_{\Phi,\,\bar\theta}\; \mathbb{E}_{i,\mathbf{x}}\!\left[\,L\!\left(f(\mathbf{x};\,\bar\theta,\,\zeta^{(i)}(\Phi)),\, s^{(i)}(\mathbf{x})\right)\,\right],
  \label{eq:amortized}
\end{equation}
where $\zeta^{(i)}(\Phi)$ is produced by the hypernetwork $T_\Phi$ in a
single forward pass (TransINR, HUVR) or recovered by the meta-learner
inner loop on a meta-learned initialization (Functa), as shown in
\cref{fig:arch}. Per-instance fitting (\cref{eq:per-instance})
generally achieves higher per-tile reconstruction quality than
amortized inference on the same tile, and we refer to the resulting
difference as the \emph{amortization gap}. Continuing the
optimization of \cref{eq:per-instance} for a single tile while
freezing $\bar\theta$ and updating only $\zeta^{(i)}$ from its amortized
initialization defines the \emph{per-instance fitting bound}, which
\cref{sec:experiments:diagnosis} uses as an analytical diagnostic.

\subsection{Method inventory}
\label{sec:method:inventory}

The four methods compared in this paper, summarized in
\cref{fig:arch}, instantiate \cref{eq:amortized} with different
realizations of the hypernetwork and different choices for the
bottleneck token $\zeta^{(i)}$ and the decoder $f_{\bar\theta}$ it
conditions. Three of them (TransINR, Functa, and HUVR) are prior
work that we reproduce; the fourth, \method{}, is introduced in
\cref{sec:method:huvr_siren}. Two interfaces for injecting $\zeta^{(i)}$
into the decoder appear: \emph{weight prediction}, where the
hypernetwork outputs the decoder's weights themselves, and
\emph{feature-wise modulation} (FiLM)~\cite{perez2018film}, where the
decoder weights are shared across tiles and the per-tile token
enters as per-layer scale and shift vectors applied to each
pre-activation; shift-only FiLM is common when the underlying decoder
is a SIREN~\cite{mehta2021modulated}.

\textbf{TransINR}~\cite{chen2022transinr} uses the monolithic
hypernetwork: a single transformer ingests the input as image
patches together with learnable weight-prediction tokens and predicts
in one forward pass the full weights of a ReLU MLP neural coordinate
decoder. There is no separate encoder, no bottleneck, and no
intermediate supervision; the per-tile tokens are the learned weight-prediction tokens themselves.

\textbf{Functa}~\cite{dupont2022functa} contains no Transformer-based
hypernetwork. It sets $\zeta^{(i)} = \boldsymbol{\varphi}^{(i)}
\in \mathbb{R}^{512}$, a single global modulation vector that
conditions every hidden layer of a shared
SIREN~\cite{sitzmann2020siren} decoder through shift-only FiLM with
per-layer linear maps. The SIREN base weights and the per-layer
modulation maps are meta-learned across the training set; at
inference, $\boldsymbol{\varphi}^{(i)}$ is recovered from a zero
initialization by $T_\text{inner}$ gradient steps on the per-tile
reconstruction loss with a meta-learned per-element learning rate.
Spatial Functa~\cite{bauer2023spatial} keeps the inner loop and
replaces the global vector with a 2D grid of latents; the rest is
structurally identical.

\textbf{HUVR}~\cite{gwilliam2026huvr} uses the factored
hypernetwork. A Vision Transformer~\cite{dosovitskiy2020vit} encoder
of the B/16 variant partitions the $256\times 256$ tile into
$P{=}256$ non-overlapping $16\times 16$-pixel patches and produces one
$768$-dimensional embedding per patch plus a global \texttt{[CLS]}
embedding; a learned linear projection compresses each embedding
from width $768$ down to a smaller \emph{bottleneck width} $D$,
giving
\begin{equation}
  \zeta^{(i)} = \big(\mathbf{c}^{(i)},\,\{\mathbf{e}^{(i,p)}\}_{p=1}^P\big) \in \mathbb{R}^{(1+P)\times D},
  \label{eq:huvr-z}
\end{equation}
with one global token $\mathbf{c}^{(i)}$ and $P$ per-patch tokens
$\mathbf{e}^{(i,p)}$. We use $D{=}32$, so
the per-tile token is $257\cdot 32 = 8{,}224$ floats. A separate
transformer decoder consumes $\zeta^{(i)}$ and outputs, per patch, a
rank-1 multiplicative modulation of the shared base weights of a
$3$-layer ReLU MLP decoder; per-patch outputs are stitched and
refined by a Conv+PixelShuffle upsampler. HUVR's training combines
heightfield reconstruction with semantic distillation from a frozen
DINOv3 teacher~\cite{simeoni2025dinov3}, which we retain unchanged.

TransINR and HUVR are therefore both feed-forward at inference,
producing $\zeta^{(i)}$ for a new tile in a single forward pass with no
per-tile optimization. Functa is the exception: the per-tile
$\boldsymbol{\varphi}^{(i)}$ must be solved for from scratch by the
meta-learner inner loop, which we account for separately when
reporting inference cost (\cref{sec:experiments:crossmethod}).

\subsection{\method{}: a domain-adapted decoder for terrain heightfields}
\label{sec:method:huvr_siren}

This paper treats a terrain heightfield as a smooth surface. The
ReLU activations in HUVR's neural coordinate decoder are
piecewise-linear, with derivatives that are piecewise-constant and
discontinuous at the activation breakpoints, and higher-order
derivatives that are identically zero almost everywhere. This is a
poor match for a surface whose downstream geomorphological
interpretation relies on continuous gradients and
Laplacians~\cite{feng2024implicitterrain}. We therefore replace the
ReLU MLP with a SIREN~\cite{sitzmann2020siren}, whose sinusoidal
activations are infinitely differentiable, so the resulting
coordinate field is smooth on $[-1,1]^2$ and its analytic gradient
$\nabla f$ and Laplacian $\nabla^2 f$ are defined everywhere.

A plain SIREN has no per-tile degree of freedom. To condition it on
the patch token we extend ModulatedSIREN's per-layer
amplitude~\cite{mehta2021modulated} with an additive shift inside
the sinusoid:
\begin{equation}
  \mathbf{h}_\ell = \boldsymbol{\alpha}_\ell^{(p)} \odot
  \sin\!\big(\omega_0\,(W_\ell\, \mathbf{h}_{\ell-1} + \mathbf{b}_\ell)
  + \boldsymbol{\varphi}_\ell^{(p)}\big),
  \label{eq:siren-amp-shift}
\end{equation}
where $\mathbf{h}_\ell$ is the layer-$\ell$ activation,
$W_\ell,\mathbf{b}_\ell$ are the layer-$\ell$ weights and bias,
$\omega_0$ is a fixed base-frequency scalar, $\odot$ is the
elementwise product, and $\boldsymbol{\alpha}_\ell^{(p)},
\boldsymbol{\varphi}_\ell^{(p)}$ are amplitude and phase-shift vectors
produced from the hypernetwork decoder's per-patch output by small
LayerNorm+linear heads. The heads are initialized so that
$\boldsymbol{\alpha}\!\approx\!\mathbf{1}$ and
$\boldsymbol{\varphi}\!\approx\!\mathbf{0}$ at the start of training,
giving a plain SIREN at initialization; the network then learns to
use both modulation channels jointly. The \method{} decoder is a
$4$-layer SIREN with hidden width $256$ and $\omega_0{=}10$, the
operating frequency matched to the $16\times 16$ per-patch sub-tile
regime; its first three layers are modulated as in
\cref{eq:siren-amp-shift} and the fourth is the unmodulated linear
output, mapping to the single-channel terrain signal.
\Cref{sec:experiments:ablation} reports robustness to $\omega_0$ and
to alternative modulation interfaces (amplitude-only, shift-only,
INCODE~\cite{kazerouni2024incode}).

Since \method{} evaluates the SIREN decoder directly at each pixel
coordinate, it no longer requires HUVR's convolutional upsampler: the
$132{,}609$-parameter SIREN body has a per-pixel decode cost of
$268$k FLOPs versus $329$k for HUVR's ReLU MLP plus
Conv+PixelShuffle, a $19\%$ reduction. Total per-tile inference cost
decreases modestly from $95.3$ to $92.0$ GFLOPs because the encoder
and hypernetwork decoder dominate. Our contribution is this bounded
domain-adapted variant of HUVR, where only the post-modulation neural
coordinate decoder is replaced and the encoder, hypernetwork decoder,
token structure, and modulation interface are kept unchanged.
\Cref{sec:experiments:crossmethod} reports the resulting benchmark
deltas across all four evaluation criteria; the
decoder-by-decoder architecture comparison is in
\cref{app:asspublished}.

\section{Experimental Results}
\label{sec:experiments}

\begin{figure*}[t]
  \centering
  \includegraphics[width=\linewidth]{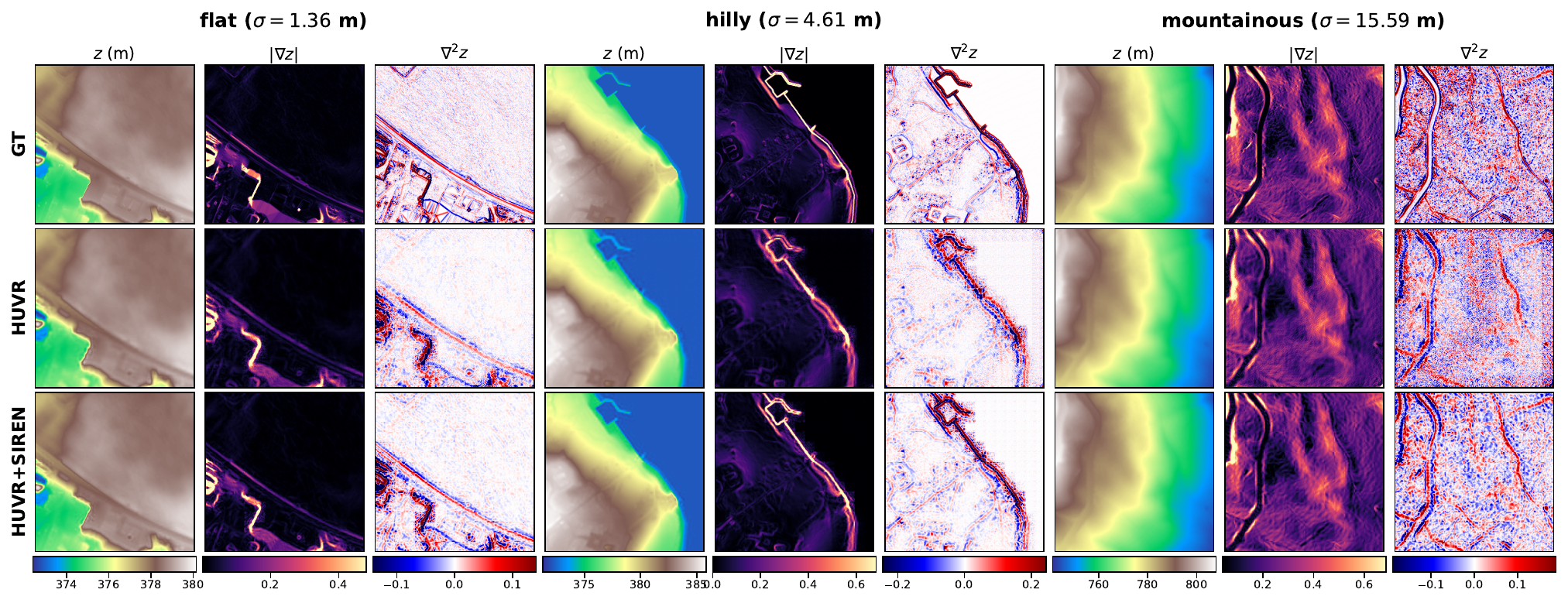}
  \caption{Reconstruction comparison on three terrain tiles selected at
    the $10\%$, $50\%$, and $90\%$ quantiles of the per-tile elevation
    standard deviation $\sigma$ (in meters), giving representative
    \emph{flat}, \emph{hilly}, and \emph{mountainous} cases. Columns
    within each tile group show elevation $z$, gradient magnitude
    $|\nabla z|$, and Laplacian $\nabla^2 z$; rows are ground truth,
    HUVR, and \method{}. Per-tile colorbars under the bottom row
    are shared by all three rows in the same tile column.}
  \Description[Three terrain tiles (flat, hilly, mountainous) shown as
    elevation, gradient magnitude, and Laplacian for ground truth,
    HUVR, and \method{}.]{A grid of imshow plots. Three column
    groups (flat, hilly, mountainous) each contain three columns
    (elevation, gradient magnitude, Laplacian). Three rows (ground
    truth, HUVR, \method{}) show that \method{} reconstructs the
    gradient and Laplacian fields more faithfully than HUVR.}
  \label{fig:grad-fidelity}
\end{figure*}

\subsection{Benchmark protocol}
\label{sec:experiments:protocol}

We benchmark on a large-scale aerial-photogrammetric digital terrain
model of Switzerland derived from swisstopo's \texttt{swissALTI3D}
(\texttt{alt3d}) product, a bare-earth elevation model that excludes
vegetation and buildings, sampled at $1\,\text{m/pixel}$. The raster
is partitioned into $256 \times 256$ tiles covering
$256\,\text{m} \times 256\,\text{m}$ each, with each tile's elevation
values normalized to zero mean and unit variance before training. The
benchmark contains $3338$ training, $420$ validation, and $432$ test
tiles. Splits are assigned at the source-region level so that no two
tiles in different splits overlap spatially. All numbers reported in
this section are computed on the held-out test split unless stated
otherwise. All experiments are conducted on a single compute node with
NVIDIA RTX A5000 GPUs.

Each method is evaluated on four criteria as a candidate terrain
data format: \emph{height fidelity}, reported as PSNR on the
normalized tile and as RMSE in meters on the de-normalized tile;
\emph{derivative fidelity}, the RMSE (in m/px and m/px$^2$ on the
de-normalized tile) of the reconstruction's gradient and Laplacian
against central finite-difference references from the ground-truth
tile, obtained analytically by automatic differentiation of the
coordinate network for the SIREN-based decoders (\method{} and Functa)
and by central finite differences of the reconstruction for the
ReLU-based decoders (TransINR and HUVR), whose activations are not
smooth; \emph{decode cost}, the analytical
FLOPs for encoding a tile and for evaluating the decoder at one
pixel coordinate; and \emph{storage}, the per-tile payload in bits
per pixel (BPP), reported both in floating-point and under
post-training quantization (PTQ) of the bottleneck token. The
detailed conventions for decode cost and the PTQ sweep are
introduced alongside the corresponding tables in
\cref{sec:experiments:crossmethod}.

For TransINR, Functa, and HUVR, the reproduction starts from the
published implementation, with a single edit to the input layer that
adapts a three-channel image-regression input to the single-channel
heightfield. TransINR uses the optimizer and learning-rate schedule
from its upstream image-regression configuration. HUVR and \method{}
use the schedule from HUVR's published code, and \method{} reuses
HUVR's encoder, patch and global tokens, and hypernetwork decoder
unchanged, differing from HUVR only in the post-modulation neural
decoder as described in \cref{sec:method:huvr_siren}. Functa is
reproduced as a meta-learned shared SIREN with $T_\text{inner}=3$
inner-loop steps at inference. Each method is trained under its own
published schedule until its validation reconstruction loss plateaus,
so the cross-method differences in \cref{sec:experiments:crossmethod}
reflect the methods rather than truncated optimization. The per-method
hyperparameter tables and the parent code revision are documented in
\cref{app:hparams}.

Two per-tile optimization diagnostics, a \emph{per-instance fitting
bound} and a \emph{per-tile full-model upper bound}, support the gap
analysis of \cref{sec:experiments:diagnosis}, where they are defined.

\subsection{Cross-method benchmark}
\label{sec:experiments:crossmethod}

We evaluate four amortized-INR methods on the swisstopo benchmark,
treating each as a candidate terrain data format and assessing it
along the four criteria defined in \cref{sec:experiments:protocol}.
\Cref{tab:headline} summarizes the result. Among the three reproduced
methods, HUVR achieves the highest height fidelity, the only
competitive derivative reconstruction, and a per-tile representation
an order of magnitude smaller than TransINR's. \method{}, our
adaptation, improves further on HUVR across every criterion we
evaluate.

\begin{table}[t]
  \centering
  % \small
  \setlength{\tabcolsep}{3pt}
  \caption{Cross-method comparison on the swisstopo terrain
    benchmark, test split ($n{=}432$). PSNR in dB,
    RMSE$_z$ in meters; $\nabla f$ (m/px) and $\nabla^2 f$ (m/px$^2$)
    are the gradient and Laplacian RMSE against finite-difference
    references, computed analytically by automatic differentiation for
    the SIREN-based decoders (HUVR+SIREN, Functa) and by central finite
    differences for the ReLU-based decoders (TransINR, HUVR).
    \emph{\#floats} is the number of scalar values in the per-tile
    payload $\zeta^{(i)}$ that must be stored for each input; the shared
    model weights are amortized across the dataset and excluded from
    the count.}
  \label{tab:headline}
  \begin{tabular}{lrrrrr}
    \toprule
    \textbf{Method} & \textbf{PSNR$\uparrow$} & \textbf{RMSE$_z\downarrow$} & \textbf{$\nabla f\downarrow$} & \textbf{${\nabla^2 f}\downarrow$} & \textbf{\#floats} \\
    \midrule
    TransINR    & 41.56 & 0.222 & 0.095 & 0.103 & 197{,}376 \\
    Functa      & 35.00 & 0.498 & 0.115 & 0.095 & 512 \\
    HUVR        & 48.86 & 0.092 & 0.068 & 0.091 & 8{,}224 \\
    \textbf{HUVR+SIREN} & \textbf{51.69} & \textbf{0.066} & \textbf{0.053} & \textbf{0.076} & 8{,}224 \\
    \bottomrule
  \end{tabular}
\end{table}

The largest single height-fidelity gain comes from the decoder
replacement. \method{} reaches $51.69\,\text{dB}$ PSNR and
$0.066\,\text{m}$ RMSE, compared with $48.86\,\text{dB}$ and
$0.092\,\text{m}$ for HUVR. This corresponds to a $+2.83\,\text{dB}$
gain and a $28\%$ reduction in height error, while using the same
encoder, hypernetwork decoder, and per-tile representation size. The
improvement also extends to derivative fidelity, with
RMSE$_{\nabla}$ reduced from $0.068$ to $0.053$ and
RMSE$_{\nabla^2}$ from $0.091$ to $0.076$. TransINR
($41.56\,\text{dB}$ / $0.222\,\text{m}$) and Functa
($35.00\,\text{dB}$ / $0.498\,\text{m}$) perform substantially below
the HUVR variants, with gaps of $7.3\,\text{dB}$ and
$13.9\,\text{dB}$ to HUVR. This indicates that the HUVR-style
patch-token representation transfers more effectively to terrain than
either full weight prediction or a single global modulation vector.
The Functa gap is also consistent with its $512$-dimensional global
modulation vector, which trades the spatially-varying structure that
HUVR's $P{=}256$ patch grid carries for a smaller per-tile payload.

Inside the HUVR family, derivative fidelity moves in the same
direction (\cref{fig:grad-fidelity} and the right half of
\cref{tab:headline}). The decoder replacement preserves the sign of
the gain at every derivative order, with the relative reduction
shrinking from $-28\%$ on height to $-23\%$ on gradient and $-17\%$
on Laplacian as the targets themselves become noisier. The
worst-tile envelope behaves the same way. The maximum gradient
error drops from $4.73$ to $3.79$ ($-20\%$) and the maximum
Laplacian error from $5.77$ to $4.67$ ($-19\%$). The separation from TransINR and
Functa is clearest on the gradient, where their RMSE is roughly twice
\method{}'s ($0.095$ for TransINR and $0.115$ for Functa against
$0.053$). On the Laplacian the gap is smaller: TransINR ($0.103$) and
Functa ($0.095$) sit above \method{}'s $0.076$ but close to the HUVR
family's $0.091$, because the second-order target is itself noisier.
The much larger height-fidelity gaps already separate these two
methods from the HUVR family, and the derivative rows do not change
the cross-method ordering.

\begin{table}[t]
  \centering
  \caption{Per-method FLOPs (GFLOPs) under the $1{:}2$
    forward-to-backward convention. Train cost is per optimizer step
    on one $256\times 256$ tile; inference is per tile. Enc.\ and
    Dec.\ are the per-tile costs of the input encoder $T_\Phi$ and of
    the hypernetwork decoder; NeuDec.\ is the neural coordinate
    decoder.}
  \label{tab:flops}
  \begin{tabular}{lrrrrr}
    \toprule
    \textbf{Method} & \textbf{Train} & \textbf{Infer} & \textbf{Enc.} & \textbf{Dec.} & \textbf{NeuDec.} \\
    \midrule
    TransINR    & 431.5   & 143.8   & 109.2 & 0.10  & 34.6 \\
    Functa      & 8{,}698.7 & 4{,}832.6 & 0.0   & 0.02  & 483.3 \\
    HUVR        & 285.8   & 95.3    & 46.6  & 27.05 & 21.6 \\
    \method{}   & 275.9   & 92.0    & 46.6  & 27.86 & 17.5 \\
    \bottomrule
  \end{tabular}
\end{table}

On the \emph{decode-cost} criterion (\cref{tab:flops}), we report
analytical FLOPs under a standard linear-layer accounting convention
with the backward pass counted at twice the forward cost. Training
cost is per optimizer step on one tile (forward plus backward) and
inference cost is per tile under a forward-only convention, with the
exception of Functa, whose meta-learning inner loop runs at inference
and is therefore counted as part of its inference budget. \method{}
is the cheapest of the four methods on both training cost per step
($275.9$ GFLOPs) and inference cost per tile ($92.0$ GFLOPs). The
per-pixel neural-decoder savings from removing HUVR's
Conv-plus-PixelShuffle upsampler (\cref{sec:method:huvr_siren}) map
to a $1.04\times$ total inference reduction, because the ViT encoder
and hypernetwork decoder dominate the per-tile pipeline. TransINR
sits at $1.56\times$ \method{} on both. Functa is the exception,
because its second-order meta-learning inner loop must be re-run at
inference time on every tile, producing a $31.5\times$ training-step
cost and a $52.5\times$ per-tile inference cost relative to
\method{}. This gap follows directly from the architecture's
inner-loop adaptation requirement at deployment. Measured per-tile
inference wall-clock on the same RTX A5000 does not track the FLOPs
count within the HUVR family. A single full decode pass takes
$15.96 \pm 0.44\,\text{ms}$ for HUVR, $18.11 \pm 0.17\,\text{ms}$ for
\method{}, and $18.31 \pm 0.18\,\text{ms}$ for TransINR ($10$ warm-up
plus $100$ timed forward passes on one test tile, batch one). \method{}
is marginally slower than HUVR in wall-clock despite its lower FLOPs
count, because evaluating the SIREN decoder independently at every one
of the $256^2$ pixel coordinates is less hardware-efficient than
HUVR's fused Conv-plus-PixelShuffle upsampler. Thus, the FLOPs reduction
from removing that upsampler does not translate into a wall-clock
reduction at this tile size.

\begin{table}[t]
  \centering
  \caption{Per-tile bits-per-pixel at fp32 and the HUVR+SIREN post-training quantization frontier. \#floats is the number of scalar values in the per-tile payload stored for each input; the shared model weights are amortized across the dataset and excluded from the count. BPP$_\mathrm{uncomp}$ is fixed-width packing; BPP$_\mathrm{ent}$ is the Shannon lower bound under per-dimension entropy coding. PSNR is in dB.}
  \label{tab:bpp}
  \begin{tabular}{lrrrr}
    \toprule
    \textbf{Method} & \textbf{\#floats} & \textbf{BPP$_\mathrm{uncomp}$} & \textbf{BPP$_\mathrm{ent}$} & \textbf{PSNR} \\
    \midrule
    Functa & 512 & 0.250 & --- & 35.00 \\
    HUVR & 8,224 & 4.016 & --- & 48.86 \\
    HUVR+SIREN & 8,224 & 4.016 & --- & 51.69 \\
    TransINR & 197,376 & 96.375 & --- & 41.56 \\
    \midrule
    \multicolumn{5}{l}{\textit{HUVR+SIREN PTQ sweep:}} \\
    \quad @ int16 & 8,224 & 2.016 & 1.009 & 51.69 \\
    \quad \textbf{@ int8} & \textbf{8,224} & \textbf{1.012} & \textbf{0.822} & \textbf{51.54} \\
    \quad @ int4 & 8,224 & 0.510 & 0.381 & 42.99 \\
    \quad @ int2 & 8,224 & 0.259 & 0.128 & 27.93 \\
    \bottomrule
  \end{tabular}
\end{table}

\begin{figure}[t]
  \centering
  \includegraphics[width=\linewidth]{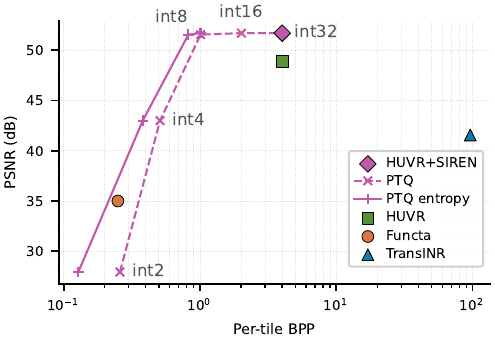}
  \caption{Rate--distortion frontier on the test split, plotting
    per-tile PSNR against per-tile bits-per-pixel under the convention
    of \cref{tab:bpp}. The \method{} PTQ frontier dominates the HUVR
    fp32 configuration down to int8.}
  \Description[Rate-distortion scatter plot showing the \method{}
    PTQ frontier dominating HUVR fp32 at three bit widths.]{
    Scatter plot of PSNR vs.\ BPP. \method{}'s fp32, int16, and int8
    points form a frontier in the upper-left that dominates HUVR's
    fp32 point. Functa fp32 sits in the lower-left at low BPP and low
    PSNR, and TransINR sits in the lower-right at high BPP and
    intermediate PSNR.}
  \label{fig:rate-distortion}
\end{figure}

Storage (\cref{tab:bpp}, \cref{fig:rate-distortion}) gives the
clearest separation. At fp32, HUVR and \method{} both cost $4.016$
BPP (the same $8{,}224$-float bottleneck token), so \method{}'s
$+2.83\,\text{dB}$ height improvement is delivered at zero additional
storage. TransINR's per-tile cost is much higher ($96.375$ BPP at
fp32, $24\times$ HUVR's bottleneck) because its transformer
specializes a $257 \times 768$ weight-token state per tile rather
than compressing through a $32$-dimensional bottleneck. Quantizing
the bottleneck token, the only source of per-tile variation in the
HUVR family, with symmetric per-dim min-max scales then exposes the
underlying compressibility of the learned representation. We say
that configuration $A$ \emph{rate-distortion dominates} configuration
$B$ when $A$ achieves no worse BPP and no worse PSNR, with a strict
improvement in at least one of the two metrics. Under this
definition, the \method{} PTQ frontier dominates the HUVR fp32
configuration for all swept bit widths down to int8. The aim of this
experiment is simply to show that \method{}'s per-tile payload
tolerates aggressive post-training quantization. Even at int8 it still
achieves a higher PSNR than HUVR's full-precision reconstruction. At
int8,
\method{} reaches $51.54\,\text{dB}$ at $0.822$ entropy-coded BPP,
reducing storage by $4.9\times$ relative to HUVR fp32 while
improving PSNR by $2.68\,\text{dB}$. Its $0.15\,\text{dB}$ loss
relative to \method{} fp32 is negligible, so we treat int8 as
preserving fp32 fidelity. Below int8 the frontier
collapses. Int4 ($0.381$ entropy-coded BPP) drops to
$42.99\,\text{dB}$, and int2 ($0.128$ entropy-coded BPP) drops to
$27.93\,\text{dB}$. Functa's fp32 configuration ($0.250$ BPP,
$35.00\,\text{dB}$) reaches a lower-BPP regime than \method{}'s
int4, but its PSNR sits more than $7\,\text{dB}$ below \method{}'s
int4 frontier point and below the fidelity regime typically required
for downstream geomorphological
analysis~\cite{feng2024implicitterrain}. Functa and \method{}
therefore occupy different rate-distortion regimes rather than
dominating each other across the full frontier.

Across the four evaluation criteria, \method{} provides the
strongest overall result among the amortized-INR methods tested on
this benchmark. Relative to HUVR, it improves height fidelity by
$+2.83\,\text{dB}$ at identical per-tile storage, reduces
derivative-fidelity error by $23\%$ to $17\%$, and gives a modest
reduction in per-tile inference cost. Its post-training
quantization frontier further dominates the HUVR fp32 operating
point down to int8, indicating that the improvement is retained
under compact storage. We therefore focus the remaining analysis on
\method{}: \cref{sec:experiments:ablation} studies the design
choices behind these gains, and
\cref{sec:experiments:diagnosis} measures the residual gap to the
per-tile fitting ceiling.

\subsection{Ablation studies}
\label{sec:experiments:ablation}

The cross-method comparison identifies \method{} as the strongest
configuration on this benchmark, but does not show which design
choices are responsible for the gain. We address that question by
varying one design choice at a time relative to the baseline configuration
($p{=}16$ pixel patches, bottleneck width $D{=}32$, SIREN base
frequency $\omega_0{=}10$, amplitude-plus-shift modulation, full
$N{=}3338$ training set with augmentation enabled). All ablation
runs share the same training recipe and evaluation protocol. The
five varied design choices are the patch size $p$ (spatial granularity at
which the encoder ingests the tile), the bottleneck width $D$
(per-tile storage capacity), the SIREN base frequency $\omega_0$
(decoder's operating frequency band), the modulation interface
(amplitude-plus-shift vs.\ amplitude-only, shift-only, and the
INCODE alternative), and the training-set size $N$ with augmentation
on or off. \Cref{tab:kaxis} summarizes the single-variable
perturbations.

\begin{table}[t]
  \centering
  \small
  \setlength{\tabcolsep}{3pt}
  \caption{One-at-a-time ablations of \method{} relative to the
    recommended configuration (top row). $\Delta$ columns are signed test-split
    differences against the baseline. $\Delta$Train reports the same
    quantity on the training split as an optimization-progress check.}
  \label{tab:kaxis}
  \begin{tabular}{lrrrrrr}
\toprule
\textbf{Config.} & \textbf{PSNR} & \textbf{$\Delta$PSNR} & \textbf{RMSE} & \textbf{$\Delta\nabla f$} & \textbf{$\Delta\nabla^{2} f$} & \textbf{$\Delta$Train} \\
\midrule
Baseline & 51.69 & --- & 0.0655 & --- & --- & --- \\
\addlinespace[2pt]
\midrule
\enspace $p{=}8$  & 56.59 & $+4.90$ & 0.0368 & $-0.0229$ & $-0.0289$ & $+4.52$ \\
\enspace $p{=}32$ & 46.65 & $-5.04$ & 0.1195 & $+0.0239$ & $+0.0162$ & $-4.94$ \\
\addlinespace[2pt]
\midrule
\enspace $D{=}16$  & 49.20 & $-2.49$ & 0.0885 & $+0.0118$ & $+0.0103$ & $-2.17$ \\
\enspace $D{=}128$ & 53.52 & $+1.82$ & 0.0529 & $-0.0085$ & $-0.0093$ & $+1.89$ \\
\enspace $D{=}256$ & 53.24 & $+1.55$ & 0.0546 & $-0.0077$ & $-0.0084$ & $+1.51$ \\
\enspace $D{=}768$ & 53.92 & $+2.22$ & 0.0505 & $-0.0104$ & $-0.0118$ & $+1.89$ \\
\addlinespace[2pt]
\midrule
\enspace $\omega_0{=}5$  & 50.78 & $-0.91$ & 0.0729 & $+0.0045$ & $+0.0043$ & $-0.92$ \\
\enspace $\omega_0{=}30$ & 51.95 & $+0.26$ & 0.0635 & $-0.0010$ & $-0.0010$ & $+0.26$ \\
\addlinespace[2pt]
\midrule
\enspace Amplitude & 51.24 & $-0.45$ & 0.0690 & $+0.0024$ & $+0.0024$ & $-0.48$ \\
\enspace Shift & 51.07 & $-0.62$ & 0.0705 & $+0.0034$ & $+0.0034$ & $-0.71$ \\
\enspace INCODE~\cite{kazerouni2024incode} & 50.33 & $-1.36$ & 0.0771 & $+0.0061$ & $+0.0053$ & $-1.12$ \\
\bottomrule
\end{tabular}

\end{table}

\paragraph{Patch size $p$.} This governs the spatial granularity at which
the encoder ingests the tile, and trades reconstruction quality
against per-tile storage. Halving $p$ from $16$ to $8$ quadruples the number of
patches $P$, so the per-tile representation size $(1+P)\cdot D$
scales as $p^{-2}$ at fixed $D$. It is also the single largest gain
we observe in the ablation, $+4.90\,\text{dB}$ at $p{=}8$ ($56.59$
vs.\ $51.69\,\text{dB}$, RMSE$_z$ $0.066 \to 0.037\,\text{m}$). Doubling
to $p{=}32$ symmetrically loses $5.04\,\text{dB}$. The gain at
smaller $p$ extends to gradient and Laplacian RMSE in the same
direction as the height RMSE, which indicates that the effect arises
from a denser spatial sampling of the input at the encoder rather
than from a better match between the per-patch neural decoder and
high-frequency content of the heightfield. We keep $p{=}16$ as the default because it provides a
better storage-quality trade-off. Using $p{=}8$ would increase the
per-tile representation from $8{,}224$ to $32{,}800$ floats and move
the model outside the storage regime analyzed in
\cref{sec:experiments:crossmethod}.

\paragraph{Patch-boundary artefacts.} Because each patch token decodes its
own sub-tile, patch-token models can introduce small height
discontinuities along patch boundaries. Every patch-token configuration we tested shows lower
PSNR within $2$ pixels of a patch seam than in the interior. The
boundary-interior gap shrinks with denser patch grids
($-1.74\,\text{dB}$ at $p{=}32$, $-1.09$ at $p{=}16$, $-0.18$ at
$p{=}8$) and with wider bottlenecks at fixed $p{=}16$ ($-1.54$ at
$D{=}16$ down to $-0.77$ at $D{=}768$). Seam artefacts are
therefore a consistent feature of patch-token architectures whose
magnitude is controlled by patch density and decoder capacity
rather than by the overall PSNR ranking. The full per-checkpoint
breakdown is in \cref{app:boundary_tokendist}.

\paragraph{Bottleneck width $D$.} This controls how much per-tile
information the encoder is allowed to keep, and is the design choice
most directly tied to per-tile storage. We sweep $D$ across $\{16, 32, 128, 256, 768\}$,
where $D{=}768$ corresponds to the no-compression upper bound at
which the encoder's $768 \to D$ projection becomes the identity.
Reconstruction improves with $D$ but with rapidly
diminishing returns. Halving the bottleneck to $D{=}16$ costs
$2.49\,\text{dB}$. Quadrupling to $D{=}128$ yields only
$+1.82\,\text{dB}$. Doubling further to $D{=}256$ changes PSNR by only
$-0.27\,\text{dB}$ from $D{=}128$, leaving it effectively
unchanged. The no-compression upper
bound at $D{=}768$ recovers $+2.22\,\text{dB}$ over $D{=}32$. The
sublinear shape of this curve shows that the $32$-dimensional
bottleneck leaves some headroom but is not the dominant constraint:
even removing the compression entirely yields only about
$+2\,\text{dB}$, less than a third of the $\approx 7\,\text{dB}$ gap to
the per-tile full-model upper bound reported in
\cref{sec:experiments:diagnosis}. We keep $D{=}32$ as the
default because the per-tile bottleneck scales linearly with
$D$. Moving to $D{=}128$ would multiply fp32 BPP from $4.016$ to
$16.06$ in exchange for $+1.82\,\text{dB}$, whereas the PTQ frontier
at $D{=}32$ already reaches $0.822$ entropy-coded BPP at
$51.54\,\text{dB}$ in \cref{sec:experiments:crossmethod}.

\paragraph{SIREN base frequency $\omega_0$.} This sets the dominant
frequency at which the neural decoder's sinusoidal activations
oscillate, and is the parameter a SIREN user typically tunes to match the
spectral content of the target signal. The natural-image SIREN default of
$\omega_0{=}30$ is calibrated to full-resolution image fitting. With
\method{}'s $16 \times 16$ per-patch regime, the effective signal
each per-patch decoder reconstructs sits in a lower frequency band,
which motivates our $\omega_0{=}10$ baseline. The ablation supports
this calibration but shows the choice is not decisive. Setting
$\omega_0{=}5$ loses $0.91\,\text{dB}$ while $\omega_0{=}30$ gains
only $0.26\,\text{dB}$. Raising $\omega_0$ above the baseline
therefore does not help, and the natural-image default of
$\omega_0{=}30$ is matched closely by our smaller-band
$\omega_0{=}10$. Only halving it to $\omega_0{=}5$ changes
reconstruction quality appreciably, by about $0.9\,\text{dB}$.

\paragraph{Modulation interface.} This determines how the hypernetwork
decoder's per-patch token influences the SIREN's pre-activation at each
layer. We compare four interfaces: amplitude-only modulation, shift-only
modulation, the combined amplitude-plus-shift modulation we use as
the baseline, and the INCODE interface~\cite{kazerouni2024incode}, a
richer modulation function introduced for natural-image fitting.
Amplitude-only and shift-only ablations lose $0.45$ and
$0.62\,\text{dB}$ respectively, indicating that both components of
the combined interface carry signal. The INCODE interface loses
$1.36\,\text{dB}$ relative to the baseline. The published gain of
INCODE on natural-image fitting therefore does not transfer to
amortized terrain reconstruction at this training-data scale. The
combined amplitude-plus-shift modulation, which is also the cheapest
of the four, remains the strongest interface we tested.

\begin{figure}[t]
  \centering
  \includegraphics[width=\linewidth]{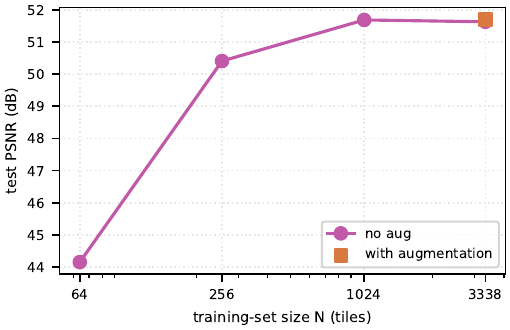}
  \caption{Dataset-size sensitivity of \method{}. Each point is a
    model trained from scratch on $N$ tiles for a fixed total of
    $21{,}000$ optimizer steps. PSNR saturates by $N \approx 1024$
    and the with-augmentation anchor at $N{=}3338$ matches the
    no-augmentation point.}
  \Description[PSNR saturation with dataset size for
    \method{}.]{PSNR (dB) plotted against $N$ on a log scale, rising
    from 44 dB at N=64 to about 52 dB at N=1024 and plateauing. The
    augmentation-on anchor at N=3338 matches the no-augmentation
    point at the same N.}
  \label{fig:data-scaling}
\end{figure}

\paragraph{Training-set size $N$ and augmentation.} This ablation
(\cref{fig:data-scaling}) tests
two related questions: how much training data the encoder-decoder
needs to amortize, and whether the HUVR augmentation policy
contributes a measurable effect on terrain. We train \method{} from
scratch on nested subsets $N \in \{64, 256, 1024, 3338\}$ of the
training split for a fixed $21{,}000$ optimizer steps without
augmentation, and re-train at the full $N{=}3338$ with augmentation
on as the comparison anchor. PSNR climbs from $44.16\,\text{dB}$ at
$N{=}64$ to $50.41\,\text{dB}$ at $N{=}256$, $51.68\,\text{dB}$ at
$N{=}1024$, and saturates at $51.62\,\text{dB}$ for the full
$N{=}3338$ split. The encoder-decoder therefore carries a real data
demand at the low end of the curve but reaches its operating regime
already by $N \approx 1024$. The final $3.3\times$ data expansion to
$N{=}3338$ contributes only $-0.06\,\text{dB}$. Enabling
augmentation at $N{=}3338$ contributes a further $+0.07\,\text{dB}$,
both negligible. Saturation at $N \approx
1024$ is consistent with the bottleneck-width result that the
$32$-dimensional bottleneck leaves only modest headroom for the
amortized representation. At this dataset scale the binding
constraint is not the supply of training data but the structural
capacity of what the shared encoder-decoder can be fit to.

\paragraph{Summary.} The decoder swap is therefore the dominant change in
\method{}, and the remaining four design choices together narrow down what
determines amortized inference quality on terrain. The
$\omega_0{=}30$ default that helps full-resolution natural-image
SIREN fitting is matched by our $\omega_0{=}10$ baseline within the
$16 \times 16$ per-patch regime. The richer INCODE modulation
interface, designed for natural-image SIREN reconstruction, does not
improve on the cheaper amplitude-plus-shift baseline at this
training-data scale. The remaining factors (bottleneck width $D$ at
fixed $p$, dataset size $N$ with or without augmentation) are
already in saturation. Building on these saturations,
\cref{sec:experiments:diagnosis} introduces two analytical diagnostics
to isolate the structural source of the residual gap to per-tile
fitting.

\subsection{Analytical diagnosis of the amortization gap}
\label{sec:experiments:diagnosis}

\method{}'s amortized inference sits at $51.69\,\text{dB}$ on the
test split, which the ablation studies of
\cref{sec:experiments:ablation} show is approximately the best
amortized number reachable by tuning the \method{} configuration
alone at this training-data scale. This subsection analyzes the
remaining gap to the per-tile fitting ceiling. We ask whether the
remaining amortization gap is mainly due to the tile-specific
bottleneck token, the shared encoder-decoder parameters, or the
structure of the learned token space. To answer this, we use two per-tile
optimization diagnostics and one structural analysis of the
bottleneck token.

\paragraph{Diagnostic construction.} The diagnostics are designed so that
each isolates one component of the pipeline as the per-tile degree of
freedom while keeping everything upstream fixed at its amortized value. The
\emph{per-instance fitting bound} freezes the hypernetwork and the
neural coordinate decoder. It then continues optimizing the bottleneck
token $\zeta^{(i)}$ alone for each tile, starting from its amortized
value. By construction it measures the
reconstruction quality the amortized encoder forgoes by producing each
tile's bottleneck token in a single forward pass rather than fitting
that token directly to the tile. A
recovery of several dB would indicate that the encoder is not using
the $32$ dimensions of the bottleneck optimally. A recovery near
zero would indicate that the encoder already produces, for each tile,
the best token the shared decoder can use. The \emph{per-tile full-model upper
bound} additionally unfreezes the hypernetwork decoder and the
SIREN base weights and jointly optimizes them with the bottleneck
token per tile, keeping only the encoder and the upstream tokenizer
frozen. By construction it measures the ceiling that this
encoder-decoder architecture could reach on each individual tile if
the shared post-bottleneck pipeline were also allowed to specialize.
This second construction is deliberately not deployable, because the whole
point of an amortized representation is that the hypernetwork
decoder and the neural decoder are shared across the dataset, so
specializing them per tile destroys the amortization advantage and
recovers an architecture that has no benefit over per-tile
optimization. We use the upper bound as a structural ceiling, not as
a proposed inference recipe.

\begin{figure}[t]
  \centering
  \includegraphics[width=\linewidth]{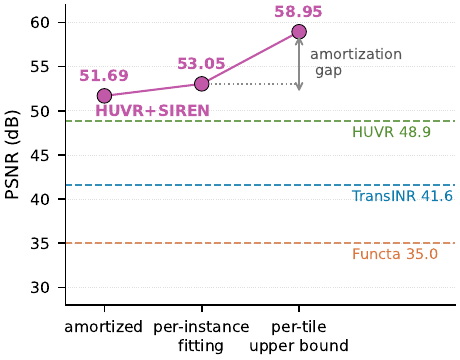}
  \caption{Decomposing the amortization gap of \method{}. The three
    red markers trace PSNR as per-tile flexibility expands: amortized
    encoding (left), the per-instance fitting bound (middle), and the
    per-tile full-model upper bound (right, computed on $32$ randomly
    sampled tiles). The vertical span on the right is the amortization
    gap and horizontal dashed lines mark the amortized cross-method
    benchmark.}
  \Description[Line plot of \method{} PSNR at three
    progressively-more-flexible per-tile configurations, with
    horizontal baselines for the other reproduced methods.]{One-column
    line plot. Y-axis: PSNR in dB. Three scatter dots from left to
    right at the amortized, per-instance-fit, and per-tile-upper-bound
    configurations of \method{}, with PSNR values 51.69, 53.05, and
    58.95. Horizontal dashed lines mark Functa at 35.00, TransINR at
    41.56, and HUVR at 48.86. The vertical span between 51.69 and
    58.95 is labeled the amortization gap.}
  \label{fig:amortization-gap}
\end{figure}

\paragraph{Decomposing the gap.} \Cref{fig:amortization-gap} reports both
diagnostics on \method{}.
The per-instance fitting bound reaches $53.05\,\text{dB}$. Continuing
to optimize the bottleneck token alone, with the shared decoder
frozen, raises amortized inference by only $1.36\,\text{dB}$. On
a per-tile basis the amortized encoder therefore already produces
patch tokens close to the best $32$-dimensional token the shared
decoder can read. The per-tile full-model upper bound, by contrast,
reaches $58.95\,\text{dB}$ after $T{=}5000$ joint-update steps on $32$
randomly sampled tiles (we sub-sample because each tile requires its
own optimization run), more than $7\,\text{dB}$ above the amortized
$51.69\,\text{dB}$ of \cref{tab:headline}. Of this roughly
$7\,\text{dB}$ gap, optimizing the bottleneck token alone closes less
than a fifth ($51.69 \to 53.05\,\text{dB}$). The remaining
$\approx 6\,\text{dB}$ does not reside in the bottleneck token but in the
shared post-bottleneck pipeline, the hypernetwork decoder and the
neural decoder whose weights are amortized across the dataset. In this
sense \method{}'s $32$-dimensional patch-token bottleneck is
approximately \emph{saturated} by amortized inference:
continuing to optimize the bottleneck token alone with the shared
decoder frozen recovers a negligible fraction of the per-tile
full-model upper bound.

\paragraph{Token-space structure.} We further characterize the bottleneck
token directly. Over all
$854{,}528$ patch tokens produced by the amortized encoder on the
training split, the PCA spectrum collapses sharply. Five of the
$32$ components account for $95\%$ of the variance and the
participation ratio (an effective-dimensionality measure equal to the
component count for a flat spectrum and tending to $1$ as variance
concentrates into a single direction; \cref{app:tokendist}) is
$1.85$. The absolute Pearson correlation coefficient $|\rho|$ between
each token dimension and the mean elevation of the patch it
represents peaks at $|\rho| = 0.072$, ruling out the trivial confound
that one or two dimensions track absolute height. The encoder
therefore concentrates per-token variance into about five effective
directions of its nominal $32$, consistent with the bottleneck-width
ablation where halving $D$ to $16$ costs only $-2.49\,\text{dB}$.
\Cref{fig:token-dist} in \cref{app:boundary_tokendist} expands this
into a four-panel diagnostic.

\paragraph{Summary.} Taken together, the two per-tile diagnostics and the
token-space analysis localize the amortization gap to the shared
components of \method{} rather than to its per-tile payload.
Re-optimizing the bottleneck token against each tile recovers only
$1.36\,\text{dB}$, so the per-tile token is already approximately
saturated. The amortized encoder also concentrates per-token variance
into roughly five of its $32$ effective directions, so the nominal
token capacity is under-used and is not the binding constraint. The remaining
$\approx 6\,\text{dB}$ of the gap to the per-tile full-model upper
bound is recovered only when the hypernetwork decoder and the SIREN
base weights are allowed to specialize, and is attributable to the
limited capacity of the shared encoder-decoder that \method{}
amortizes across the dataset. The amortization gap on terrain is thus
a property of the current hypernetwork design rather than of the
dataset scale or the width of the per-tile token, and narrowing it
requires a more expressive shared encoder and decoder or a more effective architectural design choice rather than a wider bottleneck.

\section{Conclusion}
\label{sec:conclusion}

This paper has presented the first controlled benchmark of amortized
neural representations on a high-resolution terrain DEM, introduced
\method{} as a bounded domain-adapted decoder for terrain
heightfields, and used two analytical diagnostics to localize the
residual gap between the amortized configuration and the per-tile
fitting ceiling. \method{} delivers a $+2.83\,\text{dB}$ pixel-PSNR
improvement over the strongest reproduced baseline at identical
per-tile storage, preserves the sign of the gain at every derivative
order we evaluate, and yields a post-training quantization frontier
that dominates the baseline's full-precision configuration down to
int8 inclusive. The analytical diagnostics localize the residual
amortization gap to the shared post-bottleneck pipeline rather than to
the bottleneck token, dataset scale, or modulation tuning.

The findings are bounded by the scope of the benchmark. The results
characterize a single geographic source (swisstopo \texttt{alt3d}) at
a single ground sampling distance ($1\,\text{m/pixel}$), with no
geographically held-out cross-continent evaluation and no head-to-head
comparison against production-grade DEM codecs. These
boundaries define the natural directions along which the present
characterization should be broadened in subsequent work.

Future work has two main directions. The first is to extend the
benchmark across additional geographic sources, ground sampling
distances, and production-codec baselines, so that the cross-domain
characterization begun here generalizes beyond a single national DEM.
The second is to mitigate the residual amortization gap localized in
this paper through architectural changes to the shared post-bottleneck
pipeline, with the aim of recovering more of the per-tile fitting
ceiling while using a smaller per-tile token payload than the
current $32$-dimensional patch tokens.

\bibliographystyle{ACM-Reference-Format}
\bibliography{main}

\clearpage
\appendix
\crefalias{section}{appendix}
\crefalias{subsection}{appendix}

\section{Reproduction, hyperparameters, and hardware}
\label{app:asspublished}
\label{app:hparams}

Each reproduction starts from the upstream public implementation, with
a single shared edit at the input layer that adapts a three-channel
image-regression input to the single-channel terrain heightfield.
\Cref{tab:appendix-decoder-swap} contrasts the HUVR and \method{}
neural decoders at the same encoder, hypernetwork decoder, and
$257 \times 32$ bottleneck slot. \Cref{tab:appendix-hparams} lists
per-method training schedules: TransINR follows its upstream
Imagenette image-regression recipe; Functa is a meta-learned shared
SIREN (depth $5$, width $256$, $\omega_0{=}30$) with a
$512$-dimensional global modulation vector adapted per tile by a
$T_\text{inner}{=}3$-step second-order meta-learner inner loop; HUVR
and \method{} use HUVR's published schedule with DINOv3 distillation
retained at its published weight. The per-instance fitting bound and per-tile full-model upper
bound (\cref{sec:experiments:diagnosis}) optimize per-tile state with
Adam ($\text{lr}{=}10^{-1}$ token-only, $\text{lr}{=}10^{-4}$ joint).
All jobs ran on one SLURM node with $4$~NVIDIA RTX~A5000 GPUs
($24$~GB) under PyTorch~$2.4$ DDP; per-instance fits ran on a single
A5000.

\begin{table}[h]
  \centering
  \small
  \setlength{\tabcolsep}{4pt}
  \vspace{-0.5em}
  \caption{Neural-decoder comparison between HUVR and \method{} at
    identical upstream pipeline. Per-pixel FLOPs cover the coordinate
    decoder and, for HUVR, its Conv+PixelShuffle upsampler (which
    \method{} does not use).}
  \label{tab:appendix-decoder-swap}
  \begin{tabular}{lll}
    \toprule
    & HUVR & \method{} \\
    \midrule
    Activation              & ReLU             & SIREN ($\omega_0{=}10$) \\
    Depth $\times$ width    & $3 \times 256$   & $4 \times 256$ \\
    Output convention       & Per-patch center & Per-pixel coord. \\
    Upsampler               & Conv + PixelShuffle & None \\
    Per-pixel decoder FLOPs & $329$k           & $268$k \\
    \bottomrule
  \end{tabular}
  \vspace{-0.5em}
\end{table}

\begin{table}[h]
  \centering
  \small
  \setlength{\tabcolsep}{4pt}
  \vspace{-0.5em}
  \caption{Per-method training hyperparameters.}
  \label{tab:appendix-hparams}
  \begin{tabular}{lllll}
    \toprule
    Method & Optim. & lr / sched. & Epochs & Batch \\
    \midrule
    TransINR    & Adam  & $10^{-4}$, step@80 & 1000 & 8 \\
    Functa      & Adam  & $5{\cdot}10^{-5}$, cosine & 200 & 8 \\
    HUVR        & AdamW & $10^{-4}$, cosine + wd $0.05$ & 200 & 32 \\
    \method{}   & AdamW & $10^{-4}$, cosine + wd $0.05$ & 200 & 32 \\
    \bottomrule
  \end{tabular}
  \vspace{-0.8em}
\end{table}

\begin{figure}[!htbp]
  \centering
  \vspace{-0.4em}
  \includegraphics[width=\columnwidth]{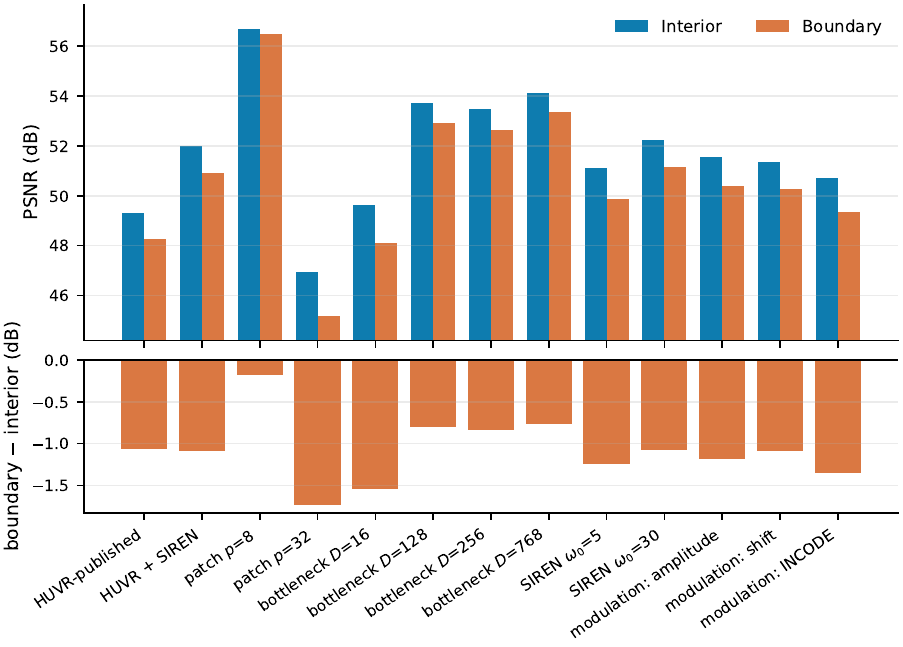}
  \vspace{-1.2em}
  \caption{Patch-boundary versus interior PSNR for every HUVR-family
    checkpoint with a patch-token bottleneck. Every point lies above
    the diagonal, so boundary PSNR is strictly below interior PSNR.}
  \Description[Scatter plot of boundary PSNR vs interior PSNR for the
    patch-token HUVR-family checkpoints.]{Scatter of boundary versus
    interior PSNR across the patch-token configurations. All points
    sit above the diagonal.}
  \label{fig:appendix-boundary}
\end{figure}

\begin{figure*}[!htbp]
  \centering
  \vspace{-0.4em}
  \includegraphics[width=0.8\linewidth]{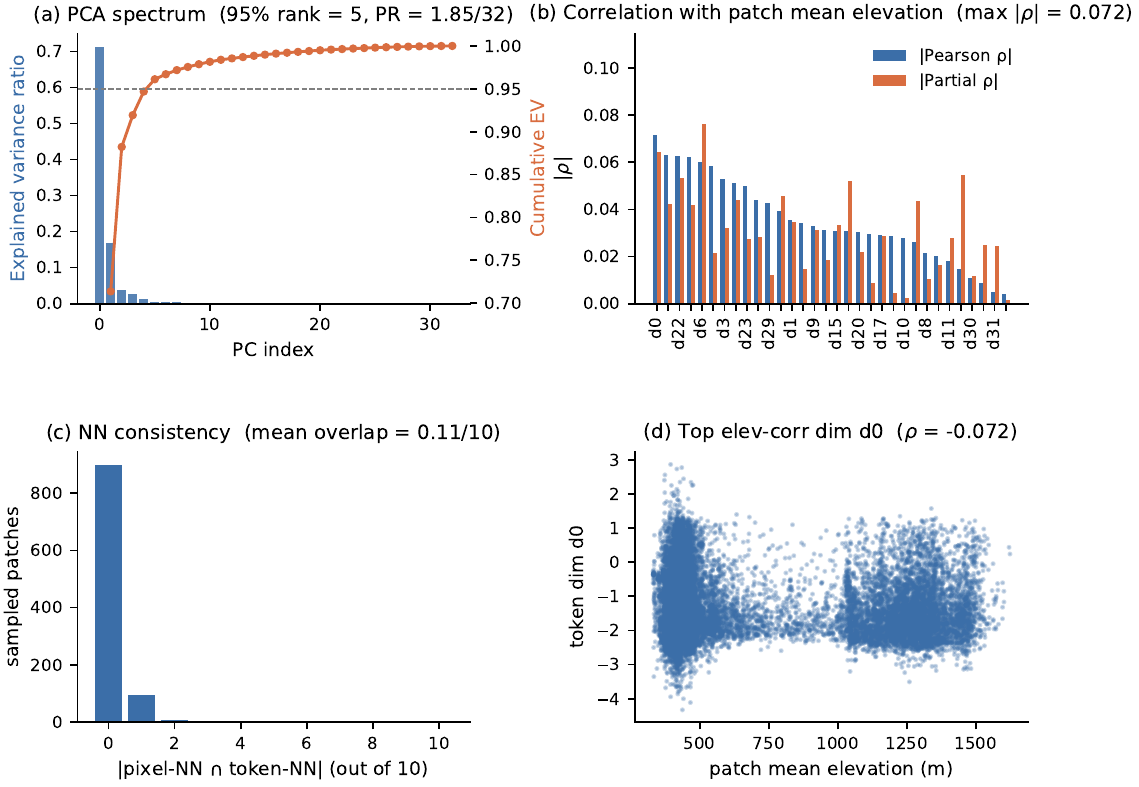}
  \vspace{-1.2em}
  \caption{Structural diagnostic of the \method{} $32$-dimensional
    patch-token bottleneck over all $854{,}528$ training-split patch
    tokens. Panels (a)--(d) defined in the surrounding text.}
  \Description[Four-panel diagnostic of the \method{} bottleneck-token
    distribution.]{Four panels labeled (a)-(d). (a) shows a
    rapidly-decaying PCA spectrum with cumulative variance crossing
    $0.95$ at component five. (b) shows that absolute Pearson and
    partial-Pearson correlations between every token dimension and
    patch mean elevation are all below $0.08$. (c) shows that the
    pixel-space and token-space $10$-nearest-neighbor sets of
    random patches share on average only $0.11$ neighbors. (d)
    shows a scatter of the most-correlated token dimension against
    patch mean elevation with no visible trend.}
  \label{fig:token-dist}
\end{figure*}

\section{Extended quantitative tables}
\label{app:extended_ablations}

\Cref{tab:appendix-gradfid} reports per-method gradient and Laplacian
RMSE on the test split together with the worst-tile error envelopes
$\max\!|\nabla\text{err}|$ and $\max\!|\nabla^2\text{err}|$ (maxima
across test tiles of the per-tile mean gradient and Laplacian error).
\method{} improves on HUVR in both summary moments and the worst-tile
envelope; the gain attenuates with derivative order, consistent with
the smooth-surface motivation of \cref{sec:method:huvr_siren}, and
the envelopes contract by a comparable factor to the
distribution-level RMSE, so the improvement is not tail-driven.
\Cref{tab:appendix-ptq} reports the \method{} PTQ frontier across
the swept bit widths $\{\text{fp32, int16, int8, int4, int2}\}$ with
both uncompressed and entropy-coded BPP under the \cref{tab:bpp}
conventions; the entropy column uses the Shannon lower bound of $32$
independent per-dimension coders fit per tile. The int8 row is the
sweet spot ($0.82$~BPP, $51.54$~dB; indistinguishable from the fp32
$51.69$~dB), while int4 collapses by $8.5$~dB and is no longer
rate-distortion-dominant.
\Cref{tab:appendix-regime} disaggregates the four-method headline by
terrain regime, with tiles binned into thirds of equal count by
HUVR's per-tile gradient-magnitude error (a ruggedness proxy uniform
across methods). PSNR is approximately regime-stable for every
method because the per-tile PSNR convention normalizes by the tile's
own elevation range; RMSE$_z$ in meters scales monotonically with
ruggedness as expected. \method{} preserves its $2$--$3$~dB gain
over HUVR within every regime and is the only method to keep
RMSE$_z$ below $0.1$~m on the rugged third.
\Cref{tab:appendix-kaxis-deriv} expands the \method{} ablation rows
of \cref{tab:kaxis} with absolute derivative RMSE on the test split;
the same ordering as \cref{tab:kaxis} is reproduced (patch size $p$
is the dominant axis, with capacity $D$ second).
\Cref{fig:appendix-boundary} reports the patch-boundary versus
interior reconstruction quality across every patch-token
configuration in the \method{} ablation studies. Every configuration
shows a strictly negative boundary-minus-interior PSNR gap, with
magnitude tracking seam density (largest at $p{=}32$, smallest at
$p{=}8$) and decoder capacity (shrinking monotonically as $D$ widens
from $16$ to $768$).

\begin{table}[h]
  \centering
  \small
  \setlength{\tabcolsep}{2pt}
  \vspace{-0.5em}
  \caption{HUVR vs.\ \method{} derivative-fidelity on the test split.
    PSNR and RMSE$_z$ repeated from \cref{tab:headline}; the last
    four columns are additional disaggregation. Gradient terms are in
    m/px and Laplacian terms in m/px$^2$.}
  \label{tab:appendix-gradfid}
  \begin{tabular}{lcccccc}
    \toprule
    Method & PSNR & RMSE$_z$ & RMSE$_{\nabla}$ & RMSE$_{\nabla^2}$ & $\max\!|\nabla\!|$ & $\max\!|\nabla^2\!|$ \\
    \midrule
    HUVR        & 48.86 & 0.092 & 0.068 & 0.091 & 4.73 & 5.77 \\
    \method{}   & 51.69 & 0.066 & 0.053 & 0.076 & 3.79 & 4.67 \\
    \bottomrule
  \end{tabular}
  \vspace{-0.8em}
\end{table}

\begin{table}[h]
  \centering
  \small
  \setlength{\tabcolsep}{4pt}
  \vspace{-0.5em}
  \caption{\method{} PTQ frontier under symmetric per-dimension
    min--max quantization with per-tile fp16 scales ($+0.008$ BPP
    overhead). BPP$_\mathrm{uncomp}$ is fixed-width packing;
    BPP$_\mathrm{ent}$ is the Shannon lower bound.}
  \label{tab:appendix-ptq}
  \begin{tabular}{lrrrr}
    \toprule
    Bit width & BPP$_\mathrm{uncomp}$ & BPP$_\mathrm{ent}$ & PSNR (dB) & RMSE (m) \\
    \midrule
    fp32  & 4.016 & ---   & 51.6907 & 0.0655 \\
    int16 & 2.016 & 1.009 & 51.6903 & 0.0655 \\
    int8  & 1.012 & 0.822 & 51.5427 & 0.0667 \\
    int4  & 0.510 & 0.381 & 42.9898 & 0.2069 \\
    int2  & 0.259 & 0.128 & 27.9307 & 1.2190 \\
    \bottomrule
  \end{tabular}
  \vspace{-0.8em}
\end{table}

\begin{table}[h]
  \centering
  \small
  \setlength{\tabcolsep}{4pt}
  \vspace{-0.5em}
  \caption{Cross-method headline by terrain regime. Tiles binned
    into thirds by HUVR's per-tile gradient-magnitude error:
    \emph{flat} (lowest), \emph{hilly}, \emph{rugged} (highest). Each
    cell is mean PSNR (dB) / RMSE$_z$ (m) over the $144$ tiles in
    that bin.}
  \label{tab:appendix-regime}
  \begin{tabular}{lccc}
    \toprule
    Method & flat & hilly & rugged \\
    \midrule
    TransINR    & 41.56\,/\,0.10 & 41.21\,/\,0.24 & 41.91\,/\,0.33 \\
    Functa      & 35.69\,/\,0.19 & 34.45\,/\,0.53 & 34.85\,/\,0.77 \\
    HUVR        & 47.96\,/\,0.04 & 49.01\,/\,0.10 & 49.61\,/\,0.14 \\
    \method{}   & 50.47\,/\,0.03 & 51.95\,/\,0.07 & 52.66\,/\,0.10 \\
    \bottomrule
  \end{tabular}
  \vspace{-0.8em}
\end{table}

\begin{table}[h]
  \centering
  \small
  \setlength{\tabcolsep}{4pt}
  \vspace{-0.5em}
  \caption{\method{} ablation rows of \cref{tab:kaxis} with absolute
    derivative RMSE on the test split.}
  \label{tab:appendix-kaxis-deriv}
  \begin{tabular}{lcc}
    \toprule
    Config. & RMSE$_{\nabla}$ (m/px) & RMSE$_{\nabla^2}$ (m/px$^2$) \\
    \midrule
    Baseline           & 0.0527 & 0.0756 \\
    $p{=}8$            & 0.0298 & 0.0467 \\
    $p{=}32$           & 0.0767 & 0.0918 \\
    $D{=}16$           & 0.0646 & 0.0858 \\
    $D{=}128$          & 0.0442 & 0.0662 \\
    $D{=}256$          & 0.0450 & 0.0671 \\
    $D{=}768$          & 0.0423 & 0.0637 \\
    $\omega_0{=}5$     & 0.0573 & 0.0799 \\
    $\omega_0{=}30$    & 0.0517 & 0.0746 \\
    amplitude          & 0.0551 & 0.0780 \\
    shift              & 0.0561 & 0.0790 \\
    INCODE             & 0.0588 & 0.0809 \\
    \bottomrule
  \end{tabular}
  \vspace{-0.8em}
\end{table}

\section{Bottleneck-token diagnostic}
\label{app:tokendist}
\label{app:boundary_tokendist}

\Cref{fig:token-dist} expands the bottleneck-token diagnostic of
\cref{sec:experiments:diagnosis} over the $854{,}528$ training-split
patch tokens. Panel (a) plots the eigenvalue spectrum
$\{\lambda_i\}_{i=1}^{32}$ of the per-token covariance matrix. Two
summaries quantify its concentration: the $95\%$-rank, the smallest
number of leading components whose cumulative variance reaches $95\%$
(here $5$), and the participation ratio $\mathrm{PR} = (\sum_i
\lambda_i)^2 / \sum_i \lambda_i^2 = 1.85$, an effective-dimensionality
measure equal to the component count for a flat spectrum (here $32$)
and tending to $1$ as variance concentrates into a single direction.
Panel (b) reports, for each of the $32$ token dimensions, its
absolute Pearson correlation coefficient $|\rho|$ with the patch mean
elevation, together with the partial-Pearson correlation that
conditions on patch elevation standard deviation and maximum gradient
norm to remove those confounds. Both stay below $0.08$, so no single
dimension encodes absolute height. Panel (c) reports
nearest-neighbor (NN) consistency, defined as the average fraction
of shared members between a patch's $k{=}10$ nearest neighbors in
raw pixel space and its $k{=}10$ nearest neighbors in token space.
Across $500$ random patches this overlap averages $0.11$ of $10$, so
token proximity does not track pixel proximity. Panel (d) scatters
the most elevation-correlated token dimension against patch mean
elevation.

\clearpage

\end{document}